# Introducing SKYSET - a Quintuple Approach for Improving Instructions

Kerry Fultz and Seth Filip

## ABSTRACT

A new approach called SKYSET (Synthetic Knowledge Yield Social Entities Translation) is proposed to validate completeness and to reduce ambiguity from written instructional documentation. SKYSET utilizes a quintuple set of standardized categories, which differs from traditional approaches that typically use triples. The SKYSET System defines the categories required to form a standard template for representing information that is portable across different domains. It provides a standardized framework that enables sentences from written instructions to be translated into sets of category typed entities on a table or database. The SKYSET entities contain conceptual units or phrases that represent information from the original source documentation. SKYSET enables information concatenation where multiple documents from different domains can be translated and combined into a single common filterable and searchable table of entities.

## 1. INTRODUCTION

Einstein famously stated *"The supreme goal of all theory is to make the irreducible basic elements as simple and as few as possible without having to surrender the adequate representation of a single datum of experience"* [3].

Instructional documentation suffers from a variety of maladies. These include a lack of commonality among processes across an enterprise, errors in documentation, ambiguity or vagueness in the way instructions were written [4], incomplete information, and complexity of directions. Communication of instructions is critical to any organized entity, whether it be a university, a medical clinic, a small business, or a large enterprise. A lack of process commonality or lack of standardization of documentation results in readers spending more time trying to make sense of the documentation. Errors, omissions, vague directions, and ambiguity in instructions potentially lead to performing tasks incorrectly [4]. This forces rework of assignments or leads to incidences that violate safety such as medical mistakes [11].

To illustrate the seriousness of mistakes, according to a 2013 report it was identified that the third leading cause of death in the United States was from hospital errors [22]. Additionally, issues with instructional complexity and lack of commonality result in documentation that is difficult to maintain. Enterprises may suffer from multiple nearly identical processes being implemented and maintained in different departments instead of using a single common process across all departments. All these issues result in risks including monetary and scheduling costs.

Even when instructional documentation does not contain errors, overhead costs still exist associated with readers spending time using the documentation. The easier documentation is to use, the less time readers need to spend using it, and the more time they can spend performing the business of the organization.

Previous approaches exist that attempt to address problems associated with written text, but these approaches have limitations and generally share similar traits. Our SKYSET (Synthetic Knowledge Yield Social Entities Translation) system is a new approach for addressing maladies encountered with the communication of written instructions. A minimal and low cost experiment was conceived and conducted with nine participants in order to perform a proof of concept for SKYSET. This paper will discuss the theory around SKYSET as well as the results of the preliminary proof of concept experiment with nine participants.

## 2. BACKGROUND AND PREVIOUS WORK

It should be noted that the SKYSET system is directly based upon the system previously called KMIE (Knowledge Management Internal External), which was created by Kerry Fultz in the 1990's. In 2013 KMIE was enhanced further and became known as SKYSET. Further SKYSET development with a proof of concept for the system's principles was performed in 2013. There has been no prior publication of SKYSET or KMIE. However, literature has been published that shares certain relevant characteristics with SKYSET and addresses problems associated with information.

### 2.1 Classical Category Concepts

Formalizations for information are not a modern innovation. Consider the following classical system in the context that it provides a framework for referencing information and explaining the world. The ancient classical Chinese system, the Wu Xing [13], incorporates five phases (Water, Wood, Fire, Earth, Metal) that are sometimes referred to as five elements, agents, processes, or movements. The Wu Xing has historically been used in association with multiple domains ranging from traditional Chinese medicine, to martial arts, to astrology.

The aspect of relevance to this discussion is that this ancient system provides a standard template of categories that may be used to reference information from multiple domains. In traditional Chinese medicine, for example, different organs in the human body are specified as belonging to specific categories. Interactions between organs and diseases may be described in context to the Wu Xing. In a different domain, such as Chinese astrology, the same five categories may be associated along with an animal on the Chinese Zodiac to specify certain attributes and interactions. This classical system is known to date back over 2000 years to at least the time of the Han dynasty [13]. It is a historical example of a categorical template applied for classification across multiple domains.

### 2.2 DNA, Building Blocks, and Language

Next consider DNA. DNA or Deoxyribonucleic acid is commonly referred to as containing the building blocks of life. A DNA double helix contains four chemical bases that encode information which are Adenine (A), Thymine (T), Cytosine (C), Guanine (G). Understanding of DNA provides insight that instructional content and complexity can be expressed through the use of a simple fundamental finite set of building blocks. Properly identifying these building blocks is critical to understanding the system.

Human language may be considered to have building blocks. Ideas expressed in language enable communication. The capability of harnessing language to label, classify, filter, and apply logic and reason with thoughts is paramount to achieving technical understanding of the world. Language behaves like a tool for thought analogous to the way a hammer is a tool for the hand. As a tool, written languages enable an expression of the building blocks of thought.

In some written languages, letters from an alphabetical system are used to represent words, while in other language systems, characters, radicals, or glyphs form words. However, phrases are constructed from sets of words, and can form written sentences. Sag, et al. [18] state that "Construction Grammarians all emphasize the importance of ... allowing units larger than the word as the building blocks of syntactic analysis". The relevance of this will be discussed in Section 3 where it will be shown that SKYSET defines a particular standard set of fundamental building blocks that adhere to specific principles and are capable of representing written information from different domains.

### 2.3 Frame Semantics and FrameNet

Frame Semantics theory was developed by Charles J. Fillmore and his colleagues [5]. It is "the study of how, as a part of our knowledge of the language, we associate linguistic forms (words, fixed phrases, grammatical patterns) with the cognitive structures – the frames". It is a theory of meaning [2] which states that "the meanings of most words can best be understood on the basis of a semantic frame... a description of a type of event, relation, or entity".

The FrameNet project, located at the International Computer Science Institute at Berkeley [2], is based upon Frame Semantics. When applied to text, FrameNet's purpose is to specify an appropriate frame, and to annotate sentences representing how a word or phrase evokes the frame. Each specified frame that is annotated is referred to as a *frame element* (FE). Example frame elements include *Sleeper*, *Avenger*, *Cook*, *Food*, *Apply_Heat*, and *Heating_instrument*. There are multitudes of frames that may exist to correspond with multitudes of sentences from different subjects.

### 2.4 Linguistics Sentence Diagramming

Sentence diagramming [8] enables a simple and generic representation of parts from a sentence into components such as Noun, Verb – Object, etc. The utility of sentence diagramming by itself is limited in part because it is too generic to provide more detailed information about the meaning of written text. This technique can, however, serve as a low level operation to assist other techniques in processing written information.

### 2.5 Controlled Natural Language (CNL)

Controlled Natural Language (CNL) [12] is a potential solution applied to instructional documentation in an effort to help reduce ambiguity and complexity. A Controlled Natural Language operates by using a subset of a natural language. In the CNL approach a restricted subset of words or grammars are permitted for use when authoring written instructions. As an example, a CNL may be applied to benefit and standardize clinical practice guidelines [21] in order to assist healthcare professionals. Due to the effort needed to write documentation that adheres to a particular CNL standard, automated grammar and style checking software tools [9] may be instituted to assist the authoring process.

### 2.6 Resource Description Framework (RDF)

The Resource Description Framework (RDF) is standard that is maintained by the RDF Working Group [16]. They describe RDF as "…a standard model for data interchange on the Web. RDF has features that facilitate data merging even if the underlying schemas differ, and it specifically supports the evolution of schemas over time without requiring all the data consumers to be changed". RDF relies upon an abstract syntax that consists of sets of "triples" [15]. Each triple consists of a subject, predicate, and object.

### 2.7 ISO 15000-5:2014 - ebXML

The International Organization for Standardization's Electronic Business Extensible Markup Language (ebXML) under ISO 15000-5:2014 defines a standardized approach to attempt to address information interoperability deficiencies between business applications. Their approach necessitates an interoperable way of standardizing Business Semantics [10]. ISO's Core Components Specification (CCS) proposes two levels of abstraction. These are Core Components and Business Information Entities.

### 2.8 Relative Abstraction of Approaches

The SKYSET system, which will be presented in Section 3, differs from previous approaches in that it uses categories containing entities with conceptual or phrase information. These SKYSET entities are at an abstraction level which is "in-between" other modalities. Hence, SKYSET operates in the "Goldilocks" zone where it is not too abstract and it is not too narrowly specific. SKYSET enables a standardized representation of written instructions from a multitude of sources. Figure 1 illustrates a notional relation between the level of abstraction and the level of restriction on languages and several associated approaches or frameworks.

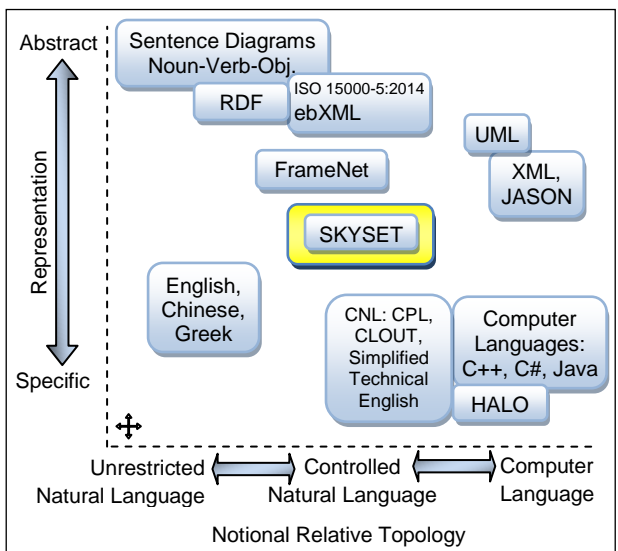

**Figure 1.** Notional level of abstraction and formalization.

In the diagram, HALO [6] refers to an "effort, sponsored by Vulcan Inc, aimed at creating the Digital Aristotle, an application that will encompass much of the world's scientific knowledge and be capable of applying sophisticated problem solving to answer novel questions." Systems like HALO may interface with CNLs.

## 3. THE SKYSET SYSTEM

SKYSET distinguishes itself from other modalities by strict adherence to a specialized set of standard categories, which are usable across multiple domains. SKYSET encompasses a language format and a translation methodology. Translation of written instructions to SKYSET format involves the use of a new process that we call Dual Intentional Semantic Decomposition and Reconstruction (DISDR). The SKYSET translation process or DISDR (pronounced *"decider"*) extracts conceptual units of information from written text that are mapped to standard categories. The resulting *SKYSET format* consists of conceptual entities that may be stored in a spreadsheet or database. SKYSET represents relationships in context between the entities.

### 3.1 SKYSET Standard Categories

SKYSET assumes that there exists a specific finite set of domain agnostic standard categories that are the building blocks of information. These are called *SKYSET Standard Categories*. These fundamental categories are presented in Table 1. Please note that we did not use standard linguistic technical terms in our SKYSET experiment and research. The SKYSET terms used in this paper have different meanings and should not be confused with technical linguistic terminology.

**Table 1.** SKYSET Standard Categories

| Category | Description | Abbreviation |
|---|---|---|
| Topic/Role | Subject of the information | TR |
| Service | Action providing or furnishing desired results for customers or products | Serv |
| Product | Something manufactured or produced | Prod |
| Resource | Noun involved in the use of or performance of services or products | Res |
| Process | Series of steps or operations toward a desired result (How to perform services or produce products) | Proc |
| Requirement | Requirements that need to be met | Req |
| Recipient (Customer) | Receiver of service or product | Recip |
| Condition (Driver) | Force or condition that drives a product or service to be delivered | Cond |

Written instructions may be translated into arrays of entities that may be represented as rows of a table, matrix, spreadsheet, or database. A complete row or array has entities that are members of each of the SKYSET Standard Categories. An entity is blank or null when a sentence lacks information pertaining to a particular category.

When applying SKYSET to documentation, some categories may be combined for simplification. An example possible quintuple with several combined categories (or table column header labels) is shown in Figure 2.

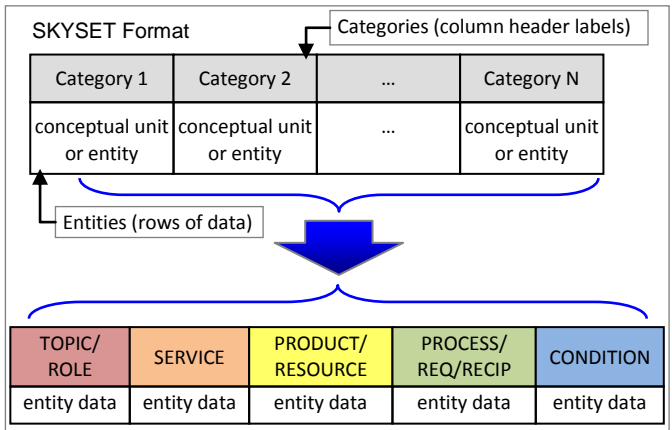

**Figure 2.** SKYSET format with categories and entities.

The categories are shown color coded. For purposes of clarity in this paper, portions of example instructional text have a color coded highlight that maps portions of the text to the appropriate corresponding color coded SKYSET category. A combined category such as Product/Resource may contain entities that belong to one or both of the respective categories.

### 3.2 SKYSET Format Introductory Example

As an introductory example, consider the following sentence shown in Example 1a.

**Example 1a.** Text to be translated into SKYSET format.

> *A scout bee reports the location of food to other scout bees via the Waggle Dance upon return to the colony*.

Dual Intentional Semantic Decomposition and Reconstruction (DISDR) may be supervised by a SKYSET practitioner (or algorithm in a computer application) to translate text into the SKYSET format. This translation of text to entities may be understood as providing answers to the following questions.

- What is the topic or role? Who is performing the action? In this text, the role is *"A Scout Bee"*.
- What action (service) is being performed? The service is *"reports"*.
- What product is being delivered, or resource used? In this case, the product is information, specifically *"the location of food"*.
- How is the service being delivered? What process is being followed or what requirement is it fulfilling? What or who is the recipient of the service and or product? In this example, the process is *"via the Waggle Dance"* and the recipient is *"to other scout bees"*.
- When or under what conditions may the action occur? The answer is *"upon return to the colony"*.

Using the DISDR process, the text *decomposes* into conceptual units or entities that map to corresponding categories. The different color highlights shown below indicate the category type that each conceptual entity maps to.

**Example 1b.** Text to be translated with highlights.

> *A scout bee* reports the location of food to other scout bees via the Waggle Dance upon return to the colony.

The individual conceptual entities may be *reconstructed* into a row of a table. Hence, this text translates to the resultant SKYSET format as shown in Table 2.

**Table 2.** SKYSET Format (for Example 1b).

| TOPIC/ ROLE | SERVICE | PRODUCT/ RESOURCE | PROCESS/ REQ/RECIPIENT | CONDITION |
|---|---|---|---|---|
| A Scout Bee | reports | the location of food | to other scout bees via the Waggle Dance | upon return to the colony |

In addition, SKYSET translation may optionally omit certain words such as definite and indefinite articles, plural or singular suffixes, etc. as deemed appropriate for the application. A simplified version is shown in Table 3.

**Table 3.** SKYSET Format (for Example 1b).

| TOPIC/ ROLE | SERVICE | PRODUCT/ RESOURCE | PROCESS/ REQ/RECIPIENT | CONDITION |
|---|---|---|---|---|
| Scout Bee | report | location of food | to other scout bees via Waggle Dance | upon return to colony |

### 3.3 SKYSET Introductory Principles

Thus far, the following SKYSET principles have been presented:

1. The DISDR process decomposes text into conceptual entities by mapping entities to SKYSET Standard Categories and assembling the entities into a row of a table. Categories are the column header labels of the table. It was observed in the course of our research that sentences rarely contain information that maps entities to every SKYSET Standard Category. When an entity belonging to a category is not represented in a sentence, it is left *blank* or has a value of *null*.

2. Two or more SKYSET Standard Categories (columns in the table) may become a set that is represented in the form of a single column. The order of columns can be rearranged.

3. If the application warrants it, certain words such as definite and indefinite articles or suffixes may be optionally omitted from the SKYSET format representation.

Additional SKYSET principles will be presented in the next several examples.

### 3.4 Ambiguity detection in text

Text is ambiguous when it contains a word, phrase, or sentence that can be interpreted to have more than one possible meaning. In their research, Zipke et al. [24] examined metalinguistic awareness involving semantic ambiguity detection. They note that metalinguistic awareness "is regarded as having special importance for helping students learn to decode words and to comprehend text".

Metalinguistic awareness, abbreviated MA, is "the ability to focus on and manipulate the formal properties of language - specifically, the ability to analyze, think about ... language as an object separate from its meaning in or out of context" [17].

Swinney et al. [14] noted in their works that meaning selection is affected by context when readers need to process sentences that contain ambiguous words. However, the research of Shakibai [20] indicate that merely perceiving a homonym is not sufficient for ambiguity perception when reading a sentence. This means detecting a word that has multiple definitions does not necessarily equate to the reader perceiving an ambiguity. Nor does it necessarily resolve a potential ambiguity or misunderstanding for the reader of the text.

Semantic ambiguities have the potential to adversely affect documentation with critical instructions interpreted to have meanings that are different from those the author of the instructions intended. SKYSET requires a formal categorization of written instructions, so ambiguities may become more readily identifiable during the translation process. In addition, mapping text appropriately to the conceptual entities potentially enables disambiguation.

The next example illustrates the property of SKYSET that enables structural ambiguity detection and resolution for written documentation.

**Example 2.** Text to be translated into SKYSET format.

*The instructor listens to the medical student with the stethoscope during class.*

This sentence can have more than one interpretation.
- **Case 1:** The instructor uses the stethoscope as a tool for examining the medical student.
- **Case 2:** The instructor hears the words spoken by the medical student. The medical student wears the stethoscope.

An alternative approach, like CNL, would necessitate replacement of words or limit the dictionary or grammar to clarify the meaning. However, unlike the CNL approach, SKYSET uses categories to identify and specify precise meaning. Using the SKYSET approach, in Case 1 the *stethoscope* is a resource used by the instructor to help perform a service. In Case 2, the recipient category is the *medical student with the stethoscope*. The corresponding possible SKYSET representations are shown in Table 8 with Case 1 and Case 2 on different rows. The author of the instructions or an appropriate knowledgeable expert would clarify which meaning needs to be represented symbolically in the written instructions using quasi-morphism logic [19]. Note that in this example, some columns swap order.

**Table 4.** SKYSET Format (for Example 2).

| Case | TOPIC/ ROLE | PRODUCT/ RESOURCE | SERVICE | PROCESS/ REQ/RECIPIENT | CONDITION |
|---|---|---|---|---|---|
| 1 | instructor | with stethoscope | listens to | medical student | during class |
| 2 | instructor | | listens to | medical student with stethoscope | during class |

As another example of ambiguous instructional text demonstrating a comparison between the SKYSET approach and a CNL approach, consider the following:

**Example 3.** Text to be translated into SKYSET format.

*The painter looks over the wall.*

Potential meanings include:
1. The painter examines the wall.
2. The painter focuses his or her visual field of view over the top of the wall to see what is on the other side of the wall.

The differences between these two meanings is important. In case 1, the instructions may need the painter to inspect the object he is working on. In case 2, the painter may need to see who or what might be on the other side of the wall before he begins spray painting. The translation to SKYSET format is shown in Table 5.

**Table 5.** SKYSET Format (for Example 3).

| Case | TOPIC/ ROLE | SERVICE | PRODUCT/ RESOURCE | PROCESS/ REQ/RECIPIENT | CONDITION |
|---|---|---|---|---|---|
| 1 | painter | looks over | the wall | | |
| 2 | painter | looks | | over the wall | |

### 3.5 Vague text

Vague written directions are too abstract to be useful in the context of the instructions.

Consider the following example. Suppose that the instructions are to train a new waiter or waitress (also known as the "server") at a formal dining establishment. The new server carries four utensils, which are a long fork, a short fork, a spoon, and a knife. The new server is completely unfamiliar with formal dining protocols, so the management creates an instructional document to guide the server step by step. An excerpt of the instructional text and the associated SKYSET format are presented below.

**Example 4.** Text to be translated into SKYSET format.

> The server places the utensil on the napkin beside the guest when the salad arrives.

**Table 6.** SKYSET Format (for Example 4).

| TOPIC/ROLE | SERVICE | PRODUCT/RESOURCE | PROCESS/REQ/RECIPIENT | CONDITION |
|---|---|---|---|---|
| server | places | utensil | on the napkin beside the guest | when the salad arrives |

Upon translation, when looking in the Product/Resource column, it becomes obvious that the word "utensil" is too vague and does not specify one of the four items (long fork, short fork, spoon, or knife) that the server carries. The knowledge that the short fork is for salads is known to the author of the instructions; however, the instructions are not explicit for someone who lacks that prior knowledge. A logical analysis of the SKYSET entities could help the author identify such issues and revise the instructions before they are published, distributed, and used. A proper revision of these directions in SKYSET format is shown below.

**Table 7.** Revision in SKYSET Format (for Example 4).

| TOPIC/ROLE | SERVICE | PRODUCT/RESOURCE | PROCESS/REQ/RECIPIENT | CONDITION |
|---|---|---|---|---|
| server | places | short fork | on the napkin beside the guest | when the salad arrives |

### 3.6 Incomplete text

Text is incomplete when required information is missing from the context of the instructions.

Consider the following standard text and associated SKYSET format.

**Example 5.** Text to be translated into SKYSET format.

> The professor should distribute the assignment before class begins.

**Table 8.** SKYSET Format (for Example 5).

| TOPIC/ROLE | SERVICE | PRODUCT/RESOURCE | PROCESS/REQ/RECIPIENT | CONDITION |
|---|---|---|---|---|
| professor | distribute | assignment | - | before class begins |

Notice that one of the categories is left blank because information about how to *distribute* the *assignment* and who to distribute it to was not specified in the instructional text. In this regard, SKYSET provides a mechanism for detection of potentially omitted instructional information. If for a given application, the process does not matter or is left up to the professor, then this omission from the instructional text may be acceptable. On the other hand, if the process is important, then the author of the instructions may identify the missing information and correct the text.

### 3.7 Many-to-One

Multiple sentences of text may translate into a single SKYSET row. When considering the nature of the framework with the standardized category style of representation, this many-to-one translation capability is a distinguishing aspect of SKYSET.

Suppose the class from the prior example text is taught virtually and the author of these instructions decides that the assignment must be distributed *by email*. Furthermore, suppose that the professor should send the instructions to the *teaching assistants* before class begin. Assume more than one sentence may be used for the information in the written instructions. The revised standard text is presented in Example 6. The equivalent translation into SKYSET format is shown in Table 9.

**Example 6.** Text to be translated into SKYSET format.

> The professor should distribute the assignment to the teaching assistants before class begins. Note that distribution should be done by email.

**Table 9.** SKYSET Format (for Example 6).

| TOPIC/ROLE | SERVICE | PRODUCT/RESOURCE | PROCESS/REQ/RECIPIENT | CONDITION |
|---|---|---|---|---|
| professor | distribute | assignment | to teaching assistants by email | before class begins |

Note that when related information from standard text spans multiple sentences, understanding it requires linking conceptual units of the information together using spatial reference in the human mind. This *information encoding* causes the information to go into units related by verbs to bring the meaning of one or more arguments into scope in order to understand the text. However, SKYSET already performs the spatial reference for the reader via the entities and their word or phrase order.

### 3.8 One-to-Many

One sentence of text may translate into several SKYSET rows. This multiple row form can be especially useful to separate out individual requirements or conditions from sentences. Consider the following standard text and associated SKYSET format shown in Example 7 and Table 10. Note that the two SKYSET rows differ only by the entities in the Condition column.

**Example 7.** Text to be translated into SKYSET format.

> In accordance with the course syllabus, the student should fill out a critique assessment form after the art is showcased in class or after a field trip to the museum.

**Table 10.** SKYSET Format (for Example 7).

| TOPIC/ ROLE | SERVICE | PRODUCT/ RESOURCE | PROCESS/ REQ/RECIPIENT | CONDITION |
|---|---|---|---|---|
| student | fill out | critique assessment form | in accordance with the course syllabus | after the art is showcased in class |
| student | fill out | critique assessment form | in accordance with the course syllabus | after a field trip to the museum |

### 3.9 Mutability

The ability of a SKYSET entity to switch association from one category to a different category depending upon the relevant context is a property that we call *mutability*. SKYSET mutability enables role switching for words, phrases and concepts within the SKYSET framework. To demonstrate mutability, consider Example 8 and the corresponding SKYSET format in Table 11. The *school library* entity changes category association depending upon the context.

**Example 8.** Text to be translated into SKYSET format.

> 1. The construction workers build the school library during the summer.
> 2. The students use the school library to study in the autumn in order to pass the exam.
> 3. The school library provides shelter during the winter.

**Table 11.** SKYSET Format (for Example 8).

| Row | TOPIC/ ROLE | SERVICE | PRODUCT/ RESOURCE | PROCESS/ REQ/RECIPIENT | CONDITION |
|---|---|---|---|---|---|
| 1 | construction workers | build | school library | – | during the summer |
| 2 | students | study | school library | in order to pass the exam | In the autumn |
| 3 | school library | provides | shelter | – | during the winter |

Notice that on Row 1, the *school library* is a product produced by the construction workers. However, when students use the *school library*, in Row 2, it becomes a resource. In Row 3, it is the topic.

### 3.10 Standardized Representation of Meaning

Another principle to note is that the SKYSET translation can be specified to favor active instead of passive voice. In agreement with Frame Semantics and the FrameNet system's concepts, we note that linguistic structure involves conceptual frames [5]. SKYSET's conceptual framework is able to present a particular standard frame for application to written instructions. A standardized representation enables sentences with identical meanings to have one common SKYSET representation.

To present this principle, the three sentences from Example 9 have identical meaning. The single corresponding SKYSET translation, shown in Table 12, is a valid translation for each of the three sentences.

**Example 9.** Text to be translated into SKYSET format.

> 1. The student fills the test tube with water when instructed.
> 2. The test tube is filled with water by the student when instructed.
> 3. When instructed, the test tube is filled with water by the student.

**Table 12.** SKYSET Format (for Example 9).

| TOPIC/ ROLE | SERVICE | PRODUCT/ RESOURCE | PROCESS/ REQ/RECIPIENT | CONDITION |
|---|---|---|---|---|
| student | fills | test tube | with water | when instructed |

Aspects of the distinguishing principles of SKYSET that have been presented thus far warrant further discussion. SKYSET enables roles that are independent which is in agreement with the cognitive linguistics style approach similar to FrameNet [2]. FrameNet creates a representation of the actual semantic frames that describe the context for which there are hundreds or thousands of potentially different semantic frames.

However, SKYSET takes a different approach from FrameNet. SKYSET postulates that it only has to assume the finite set of SKYSET Standard Categories (usually applied as a quintuple) to represent written instructions in the SKYSET format. While FrameNet has strength in its ability to capture the frame element information, it lacks a simple reusable standard template for representing categories across multiple domains due to the use of hundreds of different frame elements. In contrast, SKYSET does not attempt to capture the frame element information. Instead SKYSET focuses on capturing the conceptual entity information using the finite set of SKYSET Standard Categories as a common template that is useable across multiple domains.

Of particular relevance to the discussion of an applied system like SKYSET is an observation described by Awodey [1] in his 2015 paper. He states that "One thing missing... has been a notion of *model* that is both faithful to the precise formalism... and yet general and flexible enough to be a practical tool for semantic investigations". Hence, research has been performed and theories investigated. However, an applied approach that is practical has been elusive.

Abstract theories or generic triple approaches such as RDF have limitations in regard to their practical application. SKYSET, in contrast, defines and applies the specific SKYSET categories, typically as a quintuple template. SKYSET entities preserve information framed in context with the source instructional text. This enables the SKYSET framework to represent information at an abstraction level that facilitates practical application as a standardized format across domains.

The ISO 15000-5:2014 Core Component contains a concept that partially shares some resemblance with KMIE's/SKYSET's internal and external boundary concept that was conceived in the 1990's where business related information is included (internal) and extraneous information is excluded (external) from the KMIE/SKYSET system.

### 3.11 Information Concatenation and Integration

A key feature of SKYSET is that it identifies a finite set of categories which specify a standard template applicable to written instructions from varying subject areas. For example, the same SKYSET categories may be used to represent information from a baking instructions document and a philosophy debate team's procedural document. Furthermore, the categories are sufficiently detailed in scope as to be of enhanced value for relating information between phrasal entities. To demonstrate this, the following example is presented already in SKYSET format shown in Table 13.

**Table 13.** SKYSET Format Example

| Row | TOPIC/ ROLE | SERVICE | PRODUCT/ RESOURCE | PROCESS/ REQ/RECIPIENT | CONDITION |
|---|---|---|---|---|---|
| 1 | philosophy debate team member | wear | debate team uniform | per debate team charter | when participating in debate competition |
| 2 | baking student | choose | two baking projects | according to the course syllabus | by the third week of class |
| 3 | music major | take | intro to music class | to satisfy music department requirement | before graduation |

In this example, each row is from a different written source document: Row 1 is from a Philosophy Debate Team Charter, Row 2 is from a Baking Class Syllabus, and Row 3 is from School Graduation Requirements documentation. Suppose all three documents were translated into SKYSET format. It becomes evident that for a specific role, it would be straightforward to filter by SKYSET category or search for the subset of information that would apply directly to a particular student. If a student were part of the Philosophy debate team, planning to be a Music major and *not* taking the baking class, then only two rows would apply from the example table.

The property of SKYSET to concatenate information from many instructional source documents into a single table or database enables Information Integration. Wache, et al. [23] identify that Information Integration is necessary to contribute to addressing the interoperability problem associated with the accessibility of heterogeneous and distributed information.

### 3.12 Translation from SKYSET to Sentences

The SKYSET format may be translated back to represent the information as standard text sentences. The SKYSET format may almost be read from left to right to reconstruct a sentence with some potential word insertions or verb conjugations if applicable to the language being used. How translation back to standard sentences is performed is selected by the translator. For instance, a sentence may be translated to use active voice or passive voice as chosen by the translator. Hence, the reconstruction is tunable.

## 4. PRELIMINARY SKYSET EXPERIMENT

A preliminary SKYSET experiment was conducted with nine participants in order to obtain an initial estimation of the speed and accuracy of the new SKYSET format compared with the traditionally styled business process instructions document format. The traditional business process instructions document format is subsequently referred to as the *document format*.

### 4.1 Methodology and Procedure

Prior to the experiment, several *document format* written instructions were translated into SKYSET format. No personally identifiable information was kept on the participants. Individuals from nearby work areas volunteered their break time to participate by taking a four question exam.

Each participant was provided with a written exam. Each exam contained four questions. The length of time elapsed for a participant to attempt to answer each exam question was measured. Additionally, there was a 300 second time limit restriction for attempting to answer each question. The questions required the participants to answer using either the SKYSET format (spreadsheet containing SKYSET entities) or the *document format* (regular document style with sentences) as the source instructions.

Two of the four questions (one for SKYSET format and one for *document format*) asked the participant to find information about a single data point, such as a single word or phrase search. The other two questions (one for SKYSET format and one for *document format*) asked the participant to find information over multiple data points, where the reader needs to find and relate information about multiple words or phrases in context simultaneously in order to answer the question. The questions are identified as Q1, Q2, Q3, and Q4 and are summarized as follows.

- Q1: SKYSET format source, single data point search
- Q2: *document format* source, single data point search
- Q3: SKYSET format source, multiple data point search
- Q4: *document format* source, multiple data point search

### 4.2 Results

The experiment yielded the following results that are summarized in Table 14. The measured elapsed time to answer each question (Q1-Q4) is displayed in seconds.

**Table 14.** Exam Results Elapsed Time Data (seconds)

| Format = | SKYSET (sec) | Document (sec) | SKYSET (sec) | Document (sec) |
|---|---|---|---|---|
| | Single Data Point Search | | Multiple Data Point Search | |
| Question ID = | Q1 | Q2 | Q3 | Q4 |
| Participant 1 | 57 | 28 | 20 | 140 |
| Participant 2 | 94 | 40 | 46 | 240 |
| Participant 3 | 60 | 47 | 15 | 169 |
| Participant 4 | 71 | 52 | 40 | 143 |
| Participant 5 | 62 | 59 | 20 | 280 |
| Participant 6 | 124 | 26 | 37 | 300 |
| Participant 7 | 30 | 50 | 50 | 300 |
| Participant 8 | 75 | 48 | 92 | 137 |
| Participant 9 | 20 | 47 | 18 | 93 |
| | Single Data Point Search | | Multiple Data Point Search | |
| sum = | 593.00 | 397.00 | 338.00 | 1802.00 |
| # Samples = | 9 | 9 | 9 | 9 |
| mean = | 65.89 | 44.11 | 37.56 | 200.22 |
| std dev = | 31.22 | 10.93 | 24.25 | 80.02 |

This results indicate that a *multiple data point search* using the SKYSET format is on average over five times faster (5.3 times faster) than using the *document format*.

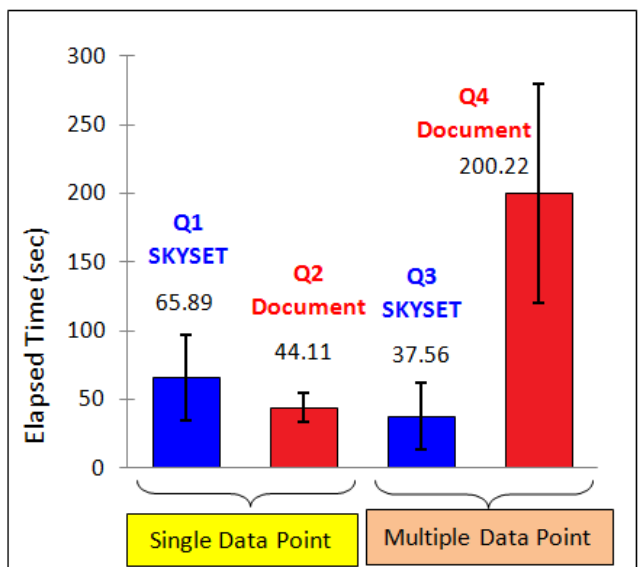

**Figure 3.** Column Chart of Average Elapsed Time (seconds) and Standard Deviation.

Figure 3 shows a chart that displays the average elapsed times in seconds and the standard deviation. It shows four columns, one for each of the four exam questions.

The left two columns represent the average elapsed time spent to answer a question requiring a *single data point search* using the SKYSET format and the *document format* respectively. The right two columns represent the average elapsed time spent to answer a question requiring a *multiple data point search* using the SKYSET format and the *document format* respectively.

There were two occurrences where test takers failed to answer a question within the time allotted because the experiment imposed a 300 second time limit per question. In both of these failures, the *document format* source material was being used. On the other hand, no failures to correctly answer a test question were observed when the source material was in SKYSET format.

## 5. DISCUSSION AND ANALYSIS

We realize that the small number of nine test participants is not sufficient to draw definitive conclusions. As discussed previously, SKYSET already provides the spatial reference for the reader. Consequently, we hypothesized that SKYSET would enable faster searching over multiple points of data. When presented with a question that requires relating multiple pieces of information, filtering or sorting by SKYSET categories enabled participants to quickly find and relate information from the SKYSET rows in order to obtain the correct answer. The observed results showed a 5.3 times speed improvement when using the SKYSET format instead of the *document format*.

Conversely, SKYSET did not show a significant difference from the *document format* when searching over a single point of data, such as using the built in search feature to locate a single word. The measured data appears to support this hypothesis. It should be emphasized again that these results are based upon only nine volunteer participants, so no definitive conclusions can be drawn. Nevertheless, it provides an initial preview of SKYSET that warrants further study to determine the validity of realizing hypothesized benefits from SKYSET.

### 5.1 Significance of the Results

To determine the statistical significance of the results, Tukey's Honest Significant Difference [7] (Tukey's HSD) test was preformed. Tukey's HSD was used to evaluate which questions are significantly different from the others. The results of Tukey's HSD test are presented in Table 15. The measured data from Question Q4 shows a significant difference from the other three questions. Moreover, questions Q1, Q2, and Q3 demonstrated no significant difference from each other. This implies that there is no measured significant difference between the SKYSET format and the *document format* for a Single Data Point search. Furthermore, these results imply that Multiple Data Point search times were measured to improve to be nearly identical to Single Data Point search times when using SKYSET.

Table 15. Tukey's HSD Test (95% confidence level)

| Quest-ions to Compare | Difference between means | Lower Confidence Interval | Upper Confidence Interval | Adjusted P Value |
|---|---|---|---|---|
| Q2-Q1 | -21.78 | -79.20 | 35.65 | 0.73 |
| Q3-Q1 | -28.33 | -85.76 | 29.09 | 0.55 |
| Q4-Q1 | 134.33 | 76.91 | 191.76 | $2.4 \times 10^{-6}$ |
| Q3-Q2 | -6.56 | -63.98 | 50.87 | 0.99 |
| Q4-Q2 | 156.11 | 98.69 | 213.53 | $1.0 \times 10^{-7}$ |
| Q4-Q3 | 162.67 | 105.24 | 220.09 | $1.0 \times 10^{-7}$ |

Within the context of the nine person experiment, sorting by SKYSET categories was observed to produce faster answer times for the SKYSET format (Q3) when compared with the times from the *document format* (Q4).

## 6. CONCLUSIONS AND FUTURE WORK

SKYSET defines a standard set of categories that are applicable as a template across multiple domains. The preliminary proof-of-concept experiment with nine participants was observed to produce an approximate 5 times speed improvement for multiple data point searches when using the SKYSET format compared with conventional instructional documentation. A speed improvement could translate into reductions in the time required to understand and follow instructions and could be indicative of clearer instructions. However, the number of experiment participants was too low to allow for any definitive conclusions. Furthermore, if this experiment were repeated with additional participants, we would need to account for the variable of the source format itself (spreadsheet versus PDF document). One potential option is to print a paper copy of the SKYSET format as a table and compare it to a paper copy of a corresponding document with sentences. Such an experiment would control for the differences between using computer applications like Excel (for SKYSET) and PDF (for the document). More studies are needed to ascertain any hypothetical benefits of SKYSET.

The observed capability of SKYSET to assist with identifying deficient documentation could help prevent mistakes that lead to malpractice or rework. SKYSET has the capability to concatenate information from many instructional source documents from different domains into a single table. Areas for future research with SKYSET are numerous. These include, but are not limited to, the following areas.

- Developing and refining efficient software algorithms that utilize DISDR to translate written information to or from the SKYSET format.

- Researching the effectiveness of SKYSET for validation of instructional documentation. This would include evaluating the effectiveness of assisting authors with detecting and resolving ambiguous, vague, or incomplete text. It could also project monetary savings by estimating the costs saved as the result of preventing malpractice or re-work.

- Research that compares SKYSET with a CNL.

- Research combining SKYSET with a CNL, where the CNL specifies the appropriate words for determining a unique representation of sentences with identical meanings.

- Researching the utility of translating text to the SKYSET format and then translating it back to a conventional document. The research could include development of parameters that are varied to produce text possessing specific stylistic traits, such as sentences with active or passive voice.

- Researching the benefits of SKYSET for information summarization and information integration. Information Summarization refers to the ability of a system to create a summary of a document's contents. Summarization is possible with SKYSET because it stores information as entities in filterable categories that can be counted and grouped to identify likely important key points from instructional text.

- Researching the capability of the SKYSET format to improve learning speed, accuracy, and knowledge retention through clearer documentation.

SKYSET has the potential to reduce ambiguity, complexity, and deficiencies associated with written instructions; however, more studies are needed to definitively prove the potential benefits. Nevertheless, the development and application of standardized formats such as SKYSET, RDF, and ISO 15000-5:2014 have the potential to foster improvements in written communication that benefit academia and other organizations.

## ACKNOWLEDGMENTS

The authors express their gratitude to all who have supported this research on its journey, including special mention to the following. Len Quadracci of Boeing Research and Technology for providing invaluable advice and support of our research. Joe Hightower, for his expertise and statistical analysis of experimental data. Gary Coen, Richard Wojcik, and Philip Harrison for their linguistics expertise and advice for the paper. Jim Troy and Brandt Dargue for their outstanding mentorship and advice.

## BIOGRAPHIES

**Kerry (Kal) Fultz** is the co-creator of SKYSET. He is a Researcher and Technical Specialist, who co-delivered the first wireless computing experience with Microsoft Windows, the early HTML knowledge base, later evolving into Microsoft Fast. Cobbled together with two others the first fully functioning version of iTunes for BSD 4.2, or native OSX for Apple, and supports Data Science currently after directly supporting the first 787-8, and 747-8's from factory to flight tests. Kal has attended University of Washington and Kaplan University.

**Seth Filip** is the co-creator of SKYSET. He is a Researcher and Engineer who hired into Boeing directly out of school. His educational background includes degrees in Physics, Electrical Engineering, and Optical Sciences, as well as certificates in Video Game Development and in 3D Virtual Worlds. Seth has experience working in Boeing Defense, Space, and Security (BDS), Boeing Research and Technology (BR&T), and Boeing Commercial Airplanes (BCA). Professional roles include Software Engineer, Data Analyst, Researcher, Instructor, Developer Lead, Virtual Collaboration Environment Expert, and Process Engineer.